\documentclass[10pt,twocolumn,letterpaper]{article}

\usepackage[pagenumbers]{cvpr} 

\usepackage[dvipsnames]{xcolor}

\definecolor{cvprblue}{rgb}{0.21,0.49,0.74}
\usepackage[pagebackref,breaklinks,colorlinks,citecolor=cvprblue]{hyperref}

\def\paperID{11216} 
\def\confName{CVPR}
\def\confYear{2024}
\usepackage{multirow}

\title{Memory-Inspired Temporal Prompt Interaction for Text-Image Classification}

\author{Xinyao Yu\\
Zhejiang University\\
{\tt\small xinyaoyu@zju.edu.cn}
\and
Hao Sun\\
Zhejiang University\\
{\tt\small sunhaoxx@zju.edu.cn}
\and
Ziwei Niu\\
Zhejiang University\\
{\tt\small nzw@zju.edu.cn}
\and
Rui Qin\\
Zhejiang University\\
{\tt\small 22260227@zju.edu.cn}
\and
Zhenjia Bai\\
Zhejiang University\\
{\tt\small 22221168@zju.edu.cn}
\and
Yen-wei Chen\\
Ritsumeikan University\\
{\tt\small chen@is.ritsumei.ac.jp}
\and
Lanfen Lin\\
Zhejiang University\\
{\tt\small llf@zju.edu.cn}
}

\begin{document}
\maketitle
\begin{abstract}
In recent years, large-scale pre-trained multimodal models (LMM) generally emerge to integrate the vision and language modalities, achieving considerable success in various natural language processing and computer vision tasks. The growing size of LMMs, however, results in a significant computational cost for fine-tuning these models for downstream tasks. Hence, prompt-based interaction strategy is studied to align modalities more efficiently. In this contex, we propose a novel prompt-based multimodal interaction strategy inspired by human memory strategy, namely \textbf{M}emory-\textbf{I}nspired \textbf{T}emporal \textbf{P}rompt Interaction (\textbf{MITP}). Our proposed method involves in two stages as in human memory strategy: the acquiring stage, and the consolidation and activation stage. We utilize temporal prompts on intermediate layers to imitate the acquiring stage, leverage similarity-based prompt interaction to imitate memory consolidation, and employ prompt generation strategy to imitate memory activation. The main strength of our paper is that we interact the prompt vectors on intermediate layers to leverage sufficient information exchange between modalities, with compressed trainable parameters and memory usage. We achieve competitive results on several datasets with relatively small memory usage and 2.0M of trainable parameters (about 1\% of the pre-trained foundation model). 
\end{abstract}

\section{Introduction}
\label{sec:intro}
Modern Internet platforms, encompassing social media and e-commerce, host a myriad of content expressed across various modalities, predominantly vision and language. Harnessing information from different modalities has demonstrated its potential to enhance performance on diverse multimodal tasks, such as image-text classification~\cite{fu2022cma}, recommendation~\cite{lu2021future}, and sentiment analysis~\cite{deng2021dense}. Multimodal learning commonly employs a strategy of interacting representations from involved modalities. Previous multimodal interaction methods~\cite{zadeh2017tensor,zadeh2018memory,hazarika2020misa,vielzeuf2018centralnet} involves extracting unimodal features separately with distinct backbones, and then blends these unimodal representations using a fusion module. However, this approach faces challenges in fully addressing comprehensive inter- and intra-modality relationships, resulting in insufficient inter-modality interaction.
 
\begin{figure*}
\centering
\begin{minipage}{0.33\linewidth}
\centering
\includegraphics[width=1\textwidth]{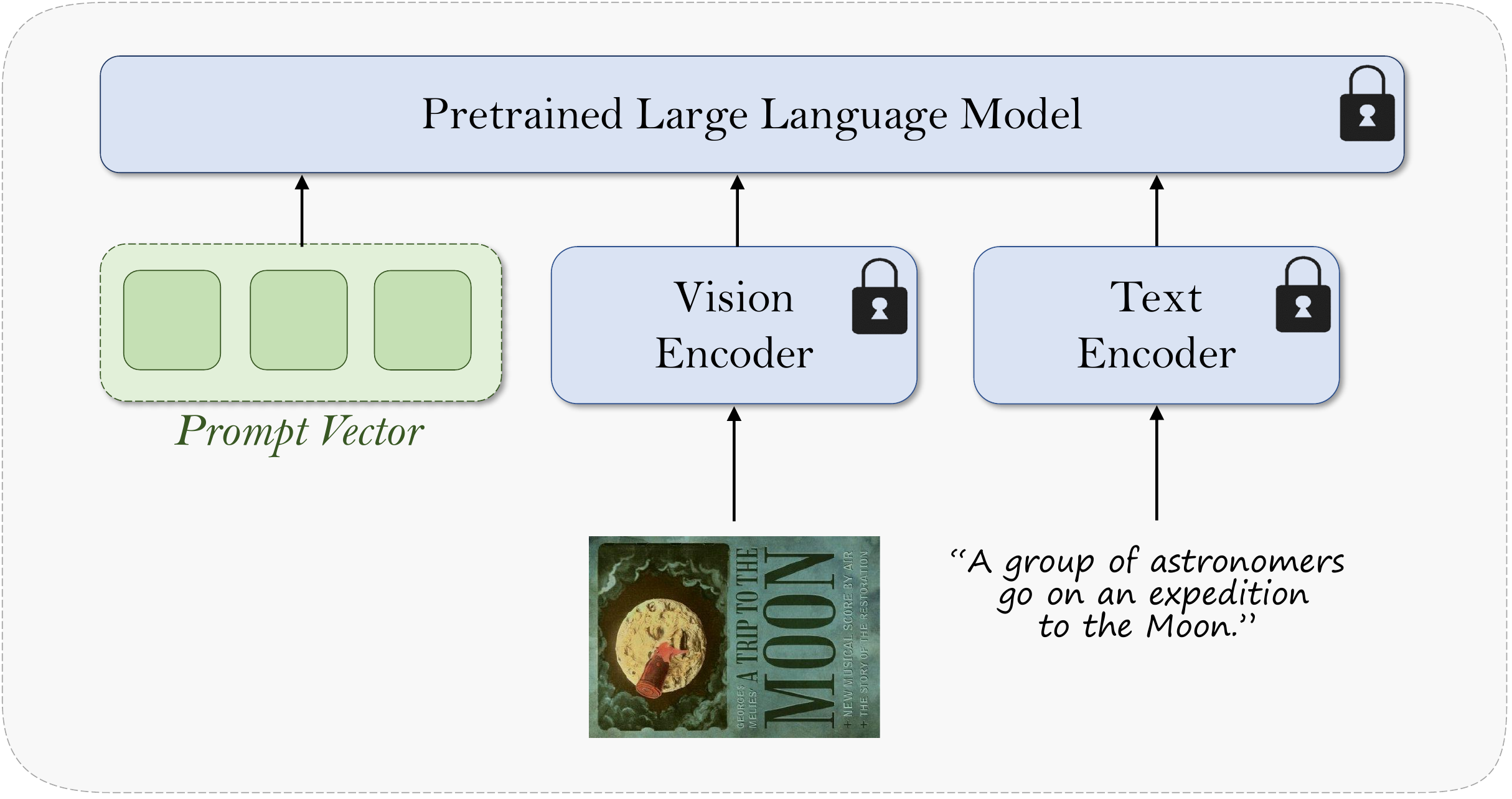}
\caption*{(a) The first proposed prompt-based interaction strategy.\cite{liang2022modular}}
\label{fig1a}
\end{minipage}
\begin{minipage}{0.33\linewidth}
\centering
\includegraphics[width=1\textwidth]{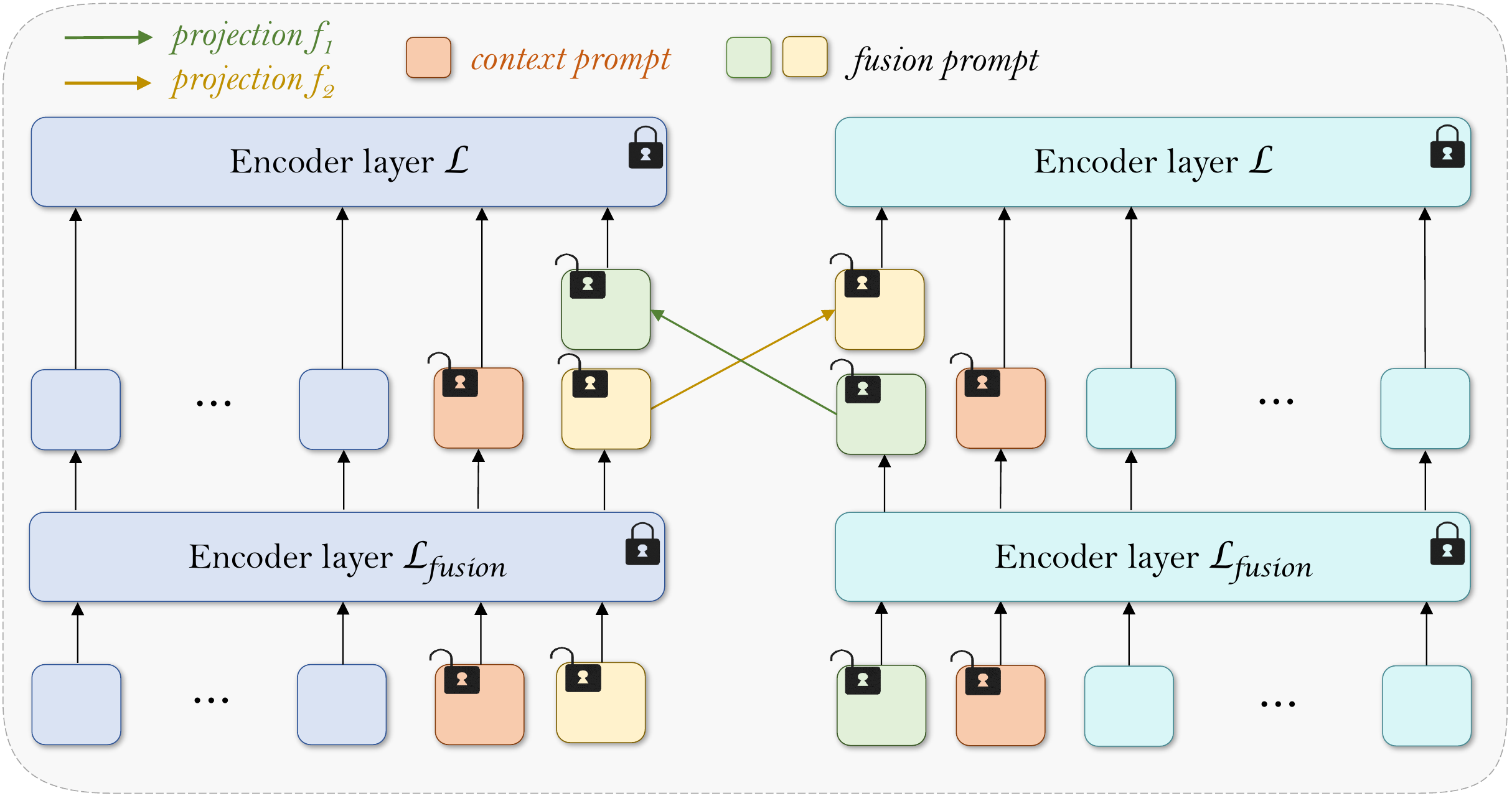}
\caption*{(b) Existing prompt-based interaction strategy.\cite{li2023efficient}}
\label{fig5b}
\end{minipage}
\begin{minipage}{0.33\linewidth}
\centering
\includegraphics[width=1\textwidth]{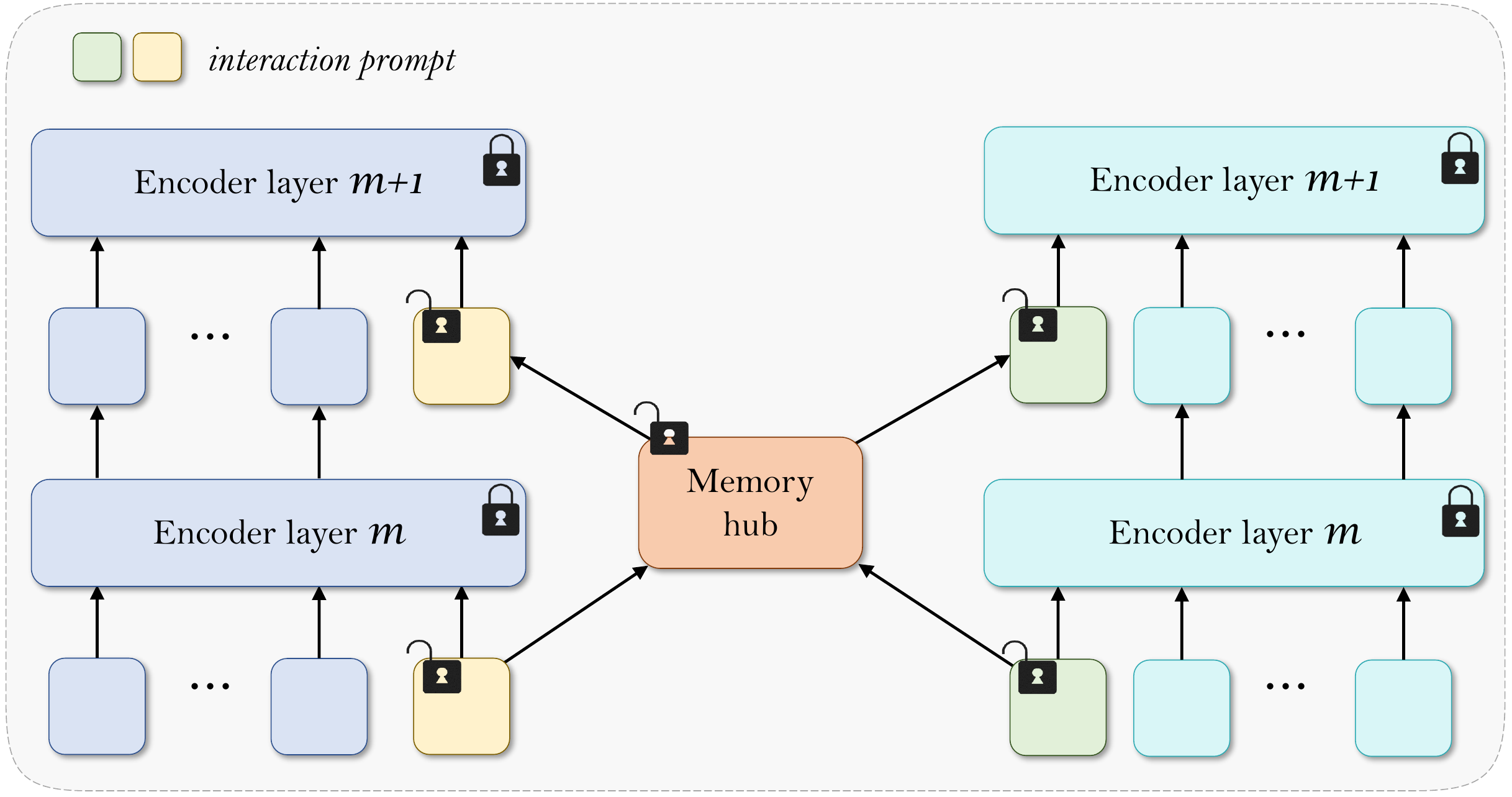}
\caption*{(c) Our proposed prompt interaction strategy MITP.}
\label{fig5c}
\end{minipage}

\caption{Comparisons among existing prompt-based interaction strategies (a)/(b), and our proposed strategy (c). (a) extracts unimodal representations separately and then utilizes prompt vectors to align the extracted representations with PLM. (b) leverages prompt vectors for each branch of the model, and utilizes the prompts to deliver information from the other branch on intermediate layers. In our proposed MITP (c), prompt vectors are leveraged for each modality on intermediate layers, and are then blended to generate prompts of the next layer. Direct interactions of prompts from different branches allow interactions between modalities.} \label{fig1}
\end{figure*}

As large-scale multimodal models (LMM) gain prominence in multimodal learning, fine-tuning methods~\cite{kiela2019supervised,jia2022visual,zhou2022learning,zhou2022conditional,khattak2023maple} for downstream tasks become prevalent, albeit with a significant drawback of large memory usage. To mitigate this issue when exploiting pre-trained models in downstream tasks, a prompt-based cross-modal interaction strategy~\cite{liang2022modular,li2023efficient} has emerged, albeit still in its infancy, as depicted in Fig~\ref{fig1} (a) and (b). The prompt-based strategy leverages prompt vectors to facilitate modality interaction on intermediate layers and mid-level features of each modality. This approach enables inter-modality information exchange through trainable prompts, reducing memory usage due to the low spatial complexity of prompt operations. However, existing methods lack direct interaction between trainable prompts, leading to insufficient interaction between modalities. The question arises: Is it possible for prompt-based interaction strategy to achieve sufficient inter-modality information exchange parameters through direct interaction of trainable prompts?

Following this trend, we propose a novel multimodal interaction method, namely \textbf{M}emory-\textbf{I}nspired \textbf{T}emporal \textbf{P}rompt Interaction (MITP). The proposed prompt-based interaction strategy is as shown in Fig~\ref{fig1} (c).
The proposed method is inspired by human memory mechanism, to be specific, working memory~\cite{baddeley2010working} and memory activation~\cite{doumas2022theory}. In working memory, information from uni-modality is acquired and registered in sensory buffers, and then consolidated into long-term memory through inter-modality interaction. Long-term memory can be activated when processing temporal information. We leverage temporal prompts on intermediate layers to acquire temporal uni-modal information, imitating the acquiring stage. After acquiring stage, temporal prompts are fed into a memory hub for prompt interaction, imitating the consolidation and activation stage. To imitate memory consolidation, we calculate similarity between temporal prompts of each modality to highlight memory logits of importance. Then, to imitate memory activation, we calculate the activation vector of prompts from each modality, and blend the activated prompts from each modality, so as to generate temporal prompts for the next layer. The interaction of temporal prompts enables a two-way information flow between both modalities without participation of modality features. Note that only the prompts and projection functions are trainable, while the rest of the model is frozen, reducing trainable parameters and memory usage significantly. Despite the variety of downstream tasks in multimodal learning, we focus on text-image classification in this paper. We achieve competitive results on several public datasets on image-text classification, such as UPMC-Food101~\cite{wang2015recipe}, MM-IMDB~\cite{arevalo2017gated}, and SNLI-VE~\cite{xie2019visual}, with only 2.0M of trainable parameters and relatively small memory usage.   

In summary, our contribution can be shown as follows:
\begin{enumerate}
    \item We propose a new form of multimodal interaction framework, namely Memory-Inspired temporal prompt interaction, which enables sufficient information flow between both modalities through direct prompt interaction.
    \item We propose temporal prompts for information storage on intermediate layers, and a memory hub for temporal prompts interaction and generating prompts for next layer, to imitate the mechanism of human memory. We further compress the number of trainable parameters by a calculative similarity-based strategy for prompt interaction, rather than a strategy with learnable parameters.
    \item Our proposed strategy significantly reduces the trainable parameters and memory usage of the model, yet we achieve competitive results on several public multimodal classification datasets. In particular, we achieve the best performance among prompt-based methods on UPMC-Food101 and SNLI-VE and surpass all other methods with the same foundation models on all three datasets.
\end{enumerate}

\begin{figure*}
    \centering
    \includegraphics[width=1\linewidth]
    {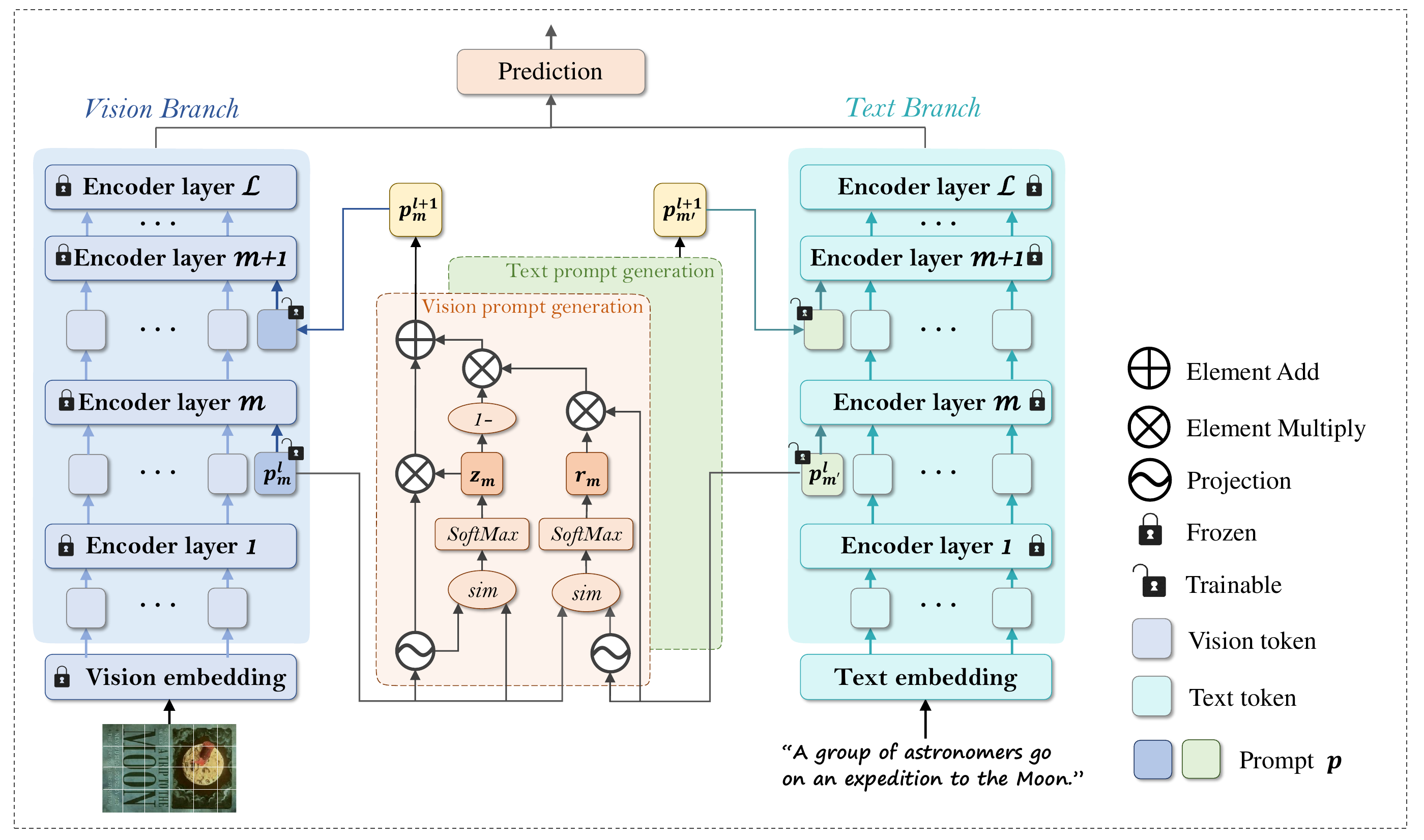}
    \caption{The pipeline of our proposed method. We utilize pre-trained foundation model in frozen for basic feature extraction of image and text branches, and leverage temporal prompts on intermediate layers to store information on temporal layer and act as media for information exchange. Temporal prompts of different modalities are blended in memory hub to generate prompts for the next layer. Only the prompts and the memory hub are trainable, requiring backward propagation; while the pre-trained foundation model of model is in frozen, which do not participate in backward propagation.}
    \label{fig2}
\end{figure*}
 

\section{Related works}
\label{sec:formatting}

In this section, we introduce previous works on multimodal interaction.

\subsection{Multimodal interaction}

Multimodal interaction aims to leverage information from different modalities for multimodal tasks. Early multimodal interaction strategy, such as TFN~\cite{zadeh2017tensor} and CentralNet~\cite{vielzeuf2018centralnet}, blends extracted features of each modality via tensor operation. Sun et al.~\cite{sun2022cubemlp} proposed CubeMLP to fuse the extracted features with an MLP-based fusion module. Although methods using tensor operation have achieved great success, the drawback of high space complexity and insufficient inter-modality interaction still exists. Therefore, attention-based methods, such as PixelBERT~\cite{huang2020pixel}, MBT~\cite{nagrani2021attention} are proposed to explore inter-modality correlation by leveraging cross modality attention to capture inter-modality correlations. Following this trend, Hazarika et al.~\cite{hazarika2020misa} proposed MISA to disentangle modality-specific and -invariant information for comprehensive and disentagled view of multimodal data. Hence, numerous works hammer at strengthening inter-modality relationship. CLMLF~\cite{li2022clmlf}, HyCon~\cite{mai2022hybrid}, and TupleInfoNCE~\cite{liu2021contrastive} utilize contrastive learning strategy to explore inter-modality, inter-class and inter-sample relationships. As LMMs become a common practice, Fu et al.~\cite{fu2022cma} tailored an attention-based fusion module for the backbone of CLIP~\cite{radford2021learning}.

\subsection{Prompt-based strategy}

Recent research has made impressive progress in large-scale multimodal pre-training~\cite{radford2021learning,liu2023gpt}. Prompt-tuning methods are proposed to finetune pre-trained large-scale multimodal models for downstream tasks. CoOp~\cite{zhou2022learning} finetunes CLIP at its language branch for few-shot transfer by a set of learnable prompt vectors, and and Co-CoOp~\cite{zhou2022conditional} promotes generalization ability of CoOp by input-conditional prompt vectors. MaPLe~\cite{khattak2023maple} extends prompt finetuning on CLIP to both textual and visual branches. In the context of the rapid growth of model size, prompt-based interaction methods are proposed to seek efficient and flexible methods other than finetuning. Liang et al.~\cite{liang2022modular} proposed PromptFuse to use prompt vectors to align the vision and language modalities, feeding prompt vectors and extracted features of each modality into PLMs as input. Li et al.\cite{li2023efficient} proposed PMF to perform multimodal interaction on intermediate layers via interacting prompts. 

This paper is in line with prompt-based interaction strategy. We store layer-wise information in temporal prompts, and perform direct interactions between prompts with memory-inspired strategy to generate temporal prompts for the next layer, to leverage in-depth interaction on mid-level features on intermediate layers.


\section{Method}
\label{method}

In this section, we present our proposed method called Memory-Inspired Temporal Prompt Interaction for image-text classification.


\subsection{Problem statement}

Given an image-text pair \(\mathit{X_{\mathit{i} } }=\left \{ \mathcal{V_{\mathit{i} }}, \mathcal{T_{\mathit{i} } }\right\}\) containing an image \(\mathcal{V_{\mathit{i} }}\in \mathbb{R^{\mathit{c\times h\times w} } } \) and a textual description \(\mathcal{T_{\mathit{i} }} \in \mathbb{R}^{\mathit{d_{i}}}\), where $c$ denotes channel, $h*w$ denotes the image size, and $d_{i}$ denotes the length of text. \(\mathit{Y_{i}} \in \{1,2,...,K\} \) denotes the class label. The goal is to learn a classifier to predict the class label \(\mathit Y_{i} \) of each image-text pair $\left\{\mathcal{V_{\mathit{i} }}, \mathcal{T_{\mathit{i} } }\right\}$.

\subsection{Model overview}

The pipeline of the model is as shown in Fig~\ref{fig2}. This work adopts LMM in frozen as foundation model to extract base features of each modality. The whole structure imitates working memory and memory activation of human memory, which can be divided into two stages: acquiring stage, and consolidation and activation stage. Temporal prompts are leveraged on several intermediate layers to acquire temporal information of this store. Then we consolidate the acquired temporal information on the basis of similarity. Finally we produce an activation mask to decide the activation rate of each logit, and add the consolidated information from the other modality, so as to generate prompts for the next layer. In practice, we select several layers to perform prompt interaction, denoted as interaction layer, while those layers without prompts are denoted as extraction layer. The consolidation and activation stages are performed in a memory hub associated to corresponding interaction layer. The selection of interaction layers and similarity type will be discussed in \ref{interaction_layers}. Note that the inter-modality interactions are conducted by prompt vectors, without participation of uni-modal features.

\subsection{Temporal prompt}
In this section, we introduce the acquiring stage. The adopted LMM foundation model contains a visual branch \(Encoder_{v}\) and a textual branch \(Encoder_{t}\). Given a pair of image and tokenized text \(\left \{ \mathcal{V_{\mathit{i} },T_{\mathit{i}} } \right\} \), we first feed the image \(\mathcal{V_{\mathit{i} }} \) into \(Encoder_{v}\) and the tokenized text \(\mathcal{T_{\mathit{i} }} \) into \(Encoder_{t}\) for basic feature extraction. The image feature and text feature on intermediate layer \(l\) are denoted as \(u_{m}^{l}\), where \(\mathit m \in \left\{v,t \right\} \) denotes modality. Temporal prompts \(\mathit{p_{m}^{l} } \) are introduced on interaction layers of each modality branch, where \(\mathit m \in \left\{v,t \right\} \) denotes modality and \(\mathit l \) refers to the interaction layer.  

Suppose this strategy starts at layer \(\mathit l_{0}\). The first temporal prompt for text \(\mathit{p_{t}^{l_{0}} } \) is initialized to be a trainable vector with Gaussion distribution. The first temporal prompt for image \(\mathit{p_{v}^{l_{0}} } \) is produced by \(\mathit{p_{t}^{l_{0}} } \) through MLP projection, as follows:

\begin{equation}
    p_{v}^{l_{0}} = W \cdot p_{t}^{l_{0}} + b \label{XX1}.
\end{equation}
On each interaction layer of branch $Encoder_{m}$, we first concatenate the modality-input \({u}_{m}^{l}  \) and the corresponding \({p_{m}^{l} } \). $p_{m}^{l}$ provides information from previous layers of both modalities, and acts as temporal memory prompt for ${u}_{m}^{l}$. Then the concatenated input is fed into the pre-trained encoder layer \(l\) as:
\begin{equation}
    [\hat{p}_{m}^{l}; {u}_{m}^{l+1} ]=Encoder_{m}^{l}([p_{m}^{l}; {u}_{m}^{l}]), \label{XX2}
\end{equation}
where \([\cdot; \cdot] \) refers to concatenation operation, and $[\hat{p}_{m}^{l}; {u}_{m}^{l+1} ]$ is the output. Through this process, information of layer $l$ from modality $m$ is acquired and registered in temporal prompt. Similarly, the input for the next layer is $[p_{m}^{l+1}; {u}_{m}^{l+1}]$. Each interaction layer is associated to a memory hub, to integrate the information from both modalities and generate temporal prompts ${p_{v}^{l+1},p_{t}^{l+1}}$ for the next layer, as:
\begin{equation}
    \left \{p_{v}^{l+1},p_{t}^{l+1} \right \} = \text{MemoryHub}(p_{v}^{l},p_{t}^{l}). \label{xx3}
\end{equation}

\subsection{Memory hub} \label{memory_hub}

In this section, we introduce the memory hub, which imitates the consolidation and activation stages of human memory. We first project $p_{m'}^{l}$ into the space of $p_{m}^{l}$ through a two-layer MLP, denoted as $\tilde p_{m'}^{l}$. In human memory consolidation, analogical mapping connection is leveraged to discover correspondences between tokens, working as $Mapping(A,B)=f_{M}(sim(A,B))$, where $f_{M}$ means the mapping function, and $sim$ means similarity function.\cite{doumas2022theory} Therefore, we mimic memory consolidation by imitating analogical mapping connection, where logits with high correspondence are to be consolidated. We consider two types of correspondences between temporal prompts: those intra-modality and those inter-modality. The correspondence mapping connections are calculated as:
\begin{equation}
    \begin{aligned}
        map_{m \gets m} & = RELU(sim(W \cdot p_{m}^{l},p_{m}^{l})), \\
        map_{m \gets m'} & = RELU(sim(p_{m}^{l}, \tilde p_{m'}^{l})),
        \label{XX4}
    \end{aligned}
\end{equation}
where $W$ denotes a two-layer MLP projection. We leverage $map_{m \gets m}$ to highlight important logits of $p_{m}^{l}$, and leverage $map_{m \gets m'}$ to highlight logits of $\tilde p_{m'}^{l}$ with high correspondence to $p_{m}^{l}$. In human memory activation, the feature unit activation is calculated by $activation_{i}=n_{i}/max(n_{j})$, where $n_{i}$ denotes feature input for unit $i$ and $max(n_{j})$ denotes the maximum feature input.\cite{doumas2022theory} In our work, activation of each logit is considered on the basis of mapping connections $map_{m \gets m}$ and $map_{m \gets m'}$, because the mapping connections represent correspondence relationships. Moreover, we consider activation in soft max manner instead of hard max. The activation of prompts is considered in intra- and inter-modality types, calculated as:
\begin{equation}
    \begin{aligned}
        z_{m} &= SoftMax(map_{m \gets m}),\\
        r_{m} &= SoftMax(map_{m \gets m'}),
        \label{XX5}
    \end{aligned}
\end{equation}
where $z_{m}$ denotes the intra-modality activation of $p_{m}^{l}$, and $r_{m}$ denotes the inter-modality activation of $\tilde p_{m'}^{l}$. Finally we activate $p_{m}^{l}$ and $\tilde p_{m'}^{l}$ to generate $p_{m}^{l+1}$, as:
\begin{equation}
    p_{m}^{l+1} = z_{m} \cdot {p}_{m}^{l}+(1-z_{m}) \cdot r_{m} \cdot \tilde {p}_{m'}^{l}.
    \label{XX6}
\end{equation}
$p_{m'}^{l+1}$ is generated by the same process as above. Through the mapping connections and activation, we can consolidate information in $p_{m}^{l}$ and integrate information from $p_{m'}^{l}$.
Note that he trainable parameters are shared in all interaction layers, so that the number of trainable parameters remains the same when the interaction layers pile up.

\subsection{Predictions}

For classification, we add hand-crafted text prompts with class labels \(y \in \left\{1,2,...,K \right\} \) having \(K\) classes. The prediction is made by the cosine similarity score with temperature \(\tau\) between the final output of the vision encoder \(x\) and the text encoder \(z\). The prediction logits \(\hat{y}=(\hat{y}_{1},...,\hat{y}_{K} )\) is calculated as:
\begin{equation}
    p(\hat{y} \mid x)=\frac{\mathrm{exp}(sim(x,z_{\hat{y}})/\tau)  }{\sum_{1}^{K} \mathrm{exp}(sim(x,z_{i})) }. 
    \label{XX7}
\end{equation}

\subsection{Loss function}

As we conduct experiments on uni-label classification task and multi-label classification task, we define loss for both tasks.
For uni-label classification task, we define the loss as:
\begin{equation}
    L_{uni} = CrossEntropy(\hat{y}_{i}, y_{i}),  \label{XX8}
\end{equation}
where \(y_{i} \in \left\{0,1 \right\}\) is the ground truth label.
For multi-label classification task, we define the loss as:
\begin{equation}
    L_{multi} = MultiCrossEntropy(\hat{y_{i}},y_{i}).    \label{XX9}
\end{equation}

\begin{table*}[]
\caption{Results on employed datasets compared with other methods. $MaPLe^{*}$ means that we re-implement MaPLe under our settings. Acc means the accuracy on UPMC-Food101 and SNLI-VE of uni-label classification task. F1-micro/F1-macro are the evaluation metrics on MM-IMDB dataset for multi-label classification task. MITP achieves competitive results and outperforms most late fusion methods as well as all prompt-based methods. Note that on SNLI-VE, we only use the hypothesis text as input for text branch, rather than both premise text and hypothesis text.}
\label{tab:my_label1}
\centering
\renewcommand\arraystretch{1.1}
\setlength{\tabcolsep}{12pt}
\begin{tabular}{lcccc}
\hline
\multirow{2}{*}{Methods}                               &  UPMC-Food101   &  MM-IMDB        &  SNLI-VE   &  Avg.\\
                                                       &  Acc(\%)        &  F1-micro/macro &  Acc($\%$) \\ \hline
 HUSE\cite{narayana2019huse}  &  92.30 & -      & -         & -\\
 VisualBERT\cite{jia2022visual}        & 92.30          & -         &   \textbf{75.06}         &\\
 Late Fusion\cite{liang2021multibench} & -          &  59.6 / 51.0             & -         & -\\
 DynMM\cite{xue2023dynamic}            & -          &  61.0 / 51.6             & -         & -\\
 UniT\cite{hu2021unit}                 & -          & -              & 73.16         & -\\
 CMA-CLIP\cite{fu2022cma}              & 93.10      &  65.3 / 52.7              & -         & -\\ \hline
 VilT\cite{pmlr-v139-kim21k}           &  92.90      & -              &  -         & -\\
 MMBT\cite{kiela2019supervised}        &  92.10      &  \textbf{66.8} / \textbf{61.8}    &  \textbf{74.69}         & \textbf{77.03}\\
 MaPLe*\cite{khattak2023maple}         & 90.80      & 60.9 / 51.2    & 71.52     & 72.79\\
 P-CLIP                                & 90.30      & 60.0 / 50.6    & 70.56  & 72.05         \\ \hline
 PromptFuse\cite{liang2022modular}     & 82.21      & 54.5 / 48.6    & 64.53        & 66.09 \\
 BlindPrompt\cite{liang2022modular}    & 84.56      & 56.5 / 50.2    & 65.54    &  67.81\\
 PMF\cite{li2023efficient}             & 91.51      & 64.5 / \textbf{58.8}    & 71.92         & 75.02\\ \hline
 MITP-base(ours)                       & \textbf{93.95} & \textbf{65.9} / 56.3         & \textbf{73.45}    & \textbf{76.17}    \\
 MITP-small(ours)                      &  93.17          & 64.8 / 55.4              & 72.51     & 75.26   \\ \hline
\end{tabular}
\end{table*}

\section{Experiments}
\label{experiments}

In this section, we introduce our datasets employed and corresponding evaluation metrics, and bring out our implementation details. 


\subsection{Datasets}
We evaluate our method on three public datasets: UPMC-Food101~\cite{wang2015recipe}, MM-IMDB~\cite{arevalo2017gated}, SNLI-VE~\cite{xie2018visual}.       
\begin{itemize}
    \item \textbf{UPMC-Food101} is a popular image-text classification dataset on food to categorize food images and its paired textual recipe descriptions into 101 categories. The dataset consists of 90,840 image-text pairs in total, including 67,971 for training and 22,715 for test.

    \item  \textbf{MM-IMDB} is a mainstream dataset for multi-label image-text classification on movie. The task is to classify the movie into one or more of the 23 genres, using the poster image and textual plot outline as image-text pairs. The dataset contains a total of 25,956 image-text pairs, with 15,510 for training, 2,599 for validation and 7,779 for test. 

    \item  \textbf{SNLI-VE} is a visual-entailment classification dataset built on the top of Flickr30K and SNLI, in which each image-text pair includes an image premise (\(P_{image}\)) and a text hypothesis (\(H_{text}\)). The task is to reason the semantic relationship (entailment, neutral, or contradiction) between \(P_{image}\) and \(H_{text}\), where \(entailment\) stands if there is enough evidence in \(P_{image}\) to deduct \(H_{text}\) as True, \(contradiction\) holds if enough evidence exists in \(P_{image}\) implying \(H_{text}\) to be False; Otherwise if the relationship is \(neutral\), it means that the evidence in \(P_{image}\) is insufficient to draw a conclusion about \(H_{text}\). The dataset has 565,286 image-text pairs in total, with 529,527 for training, 17,858 for validation and 17,901 for test. In this work, we only use the hypothesis description as text input, in line with prior work~\cite{li2023efficient}.
\end{itemize}

\subsection{Implementation details and evaluation metrics}

\begin{itemize}
    \item \textbf{Pre-trained foundation model and initialization}. We use pre-trained CLIP \(ViT-L/14\)\cite{radford2021learning} as foundation model for image encoder and text encoder in frozen. The foundation model model is pre-trained on 400 billion WIT data.  Our first temporal prompt for text branch is initialized through a Gaussion distribution (\(mean=0, std=0.02)\)), by which other prompts are generated.
    \item \textbf{Network training}. We choose Adam optimizer in all experiments, with learning rate set to 0.0002. The batch size (\(bs\)) is set to 10 for UPMC-Food101, 32 for MM-IMDB, and 64 for SNLI-VE. All our experiments are conducted on one NVIDIA RTX A6000 GPU card. Our approach is implemented in the PyTorch framework.
    \item \textbf{Metrics.} We employ accuracy\( (\%) \) as metric for uni-label classification task on UPMC-Food101 and SNLI-VE datasets, and F1-micro/F1-macro as metric for multi-label classification task on MM-IMDB dataset, following previous works~\cite{fu2022cma,li2023efficient,xue2023dynamic}.
    \item \textbf{Interaction settings.} We set the length of prompt vectors to 3 for UPMC-Food101, 6 for SNLI-VE, and 2 for MM-IMDB. We leverage 3 interaction layers empirically for the base MITP. We also propose a small version of the proposed method, utilizing one interaction layer, namely MITP-small. The selection of interaction layers will be discussed in \ref{interaction_layers}.  
\end{itemize}

\section{Results and discussion}
In this section, we report the performance of several baselines and analyze the performance of our proposed MITP on public image-text classification datasets, UPMC-Food101, MM-IMDB, and SNLI-VE.

\subsection{Baselines}

We implement several baseline models on our employed datasets and report the results for comparison. Our baseline models are as follows:

\textbf{Pre-trained LMM foundation model.} We re-implement the employed pre-trained foundation model with a linear probe under our settings and report the results. 

\textbf{Pre-trained LMM foundation model with late fusion module.} CMA-CLIP utilizes the same foundation model as our proposed MITP, complementing unimodal representations with an attention-based late fusion module. We report the results of CMA-CLIP for comparison between our proposed MITP and late fusion strategy.

\textbf{Pre-trained LMM foundation model with prompt-tuning.} MaPLe~\cite{khattak2023maple} is a multimodal prompt-tuning method on the same LMM foundation model as ours, utilizing a one-way projection from textual-branch prompts to vision-branch prompts. We re-implement MaPLe under our experimental settings on our employed image-text classification datasets and report the results. We also propose a prompt-tuning CLIP without interactions between modalities, denoted as P-CLIP, to compare our proposed method with prompt-tuning strategy. In P-CLIP, we implement deep prompt-tuning on both vision and text branches, with the pre-trained foundation model in frozen. All the prompt vectors are initialized through a Gaussion distribution ($mean=0, std=0.02$). We analysis the efficiency of trainable parameters and memory usage of both prompt-tuning methods.

\textbf{Existing prompt-based methods for multimodal interaction.} PromptFuse~\cite{liang2022modular}, BlindPrompt~\cite{liang2022modular}, and PMF~\cite{li2023efficient} are existing prompt-based interaction methods. We compare the results on employed datasets, and analyze the efficiency of trainable parameters and memory usage.

\textbf{Late fusion methods with a fusion module.} \cite{narayana2019huse,liang2021multibench,hu2021unit,li1908visualbert,xue2023dynamic} are late fusion methods, we report the results of them for comparison.

\begin{figure*}
\centering
\begin{minipage}{0.27\linewidth}
\centering
\includegraphics[width=1\textwidth]{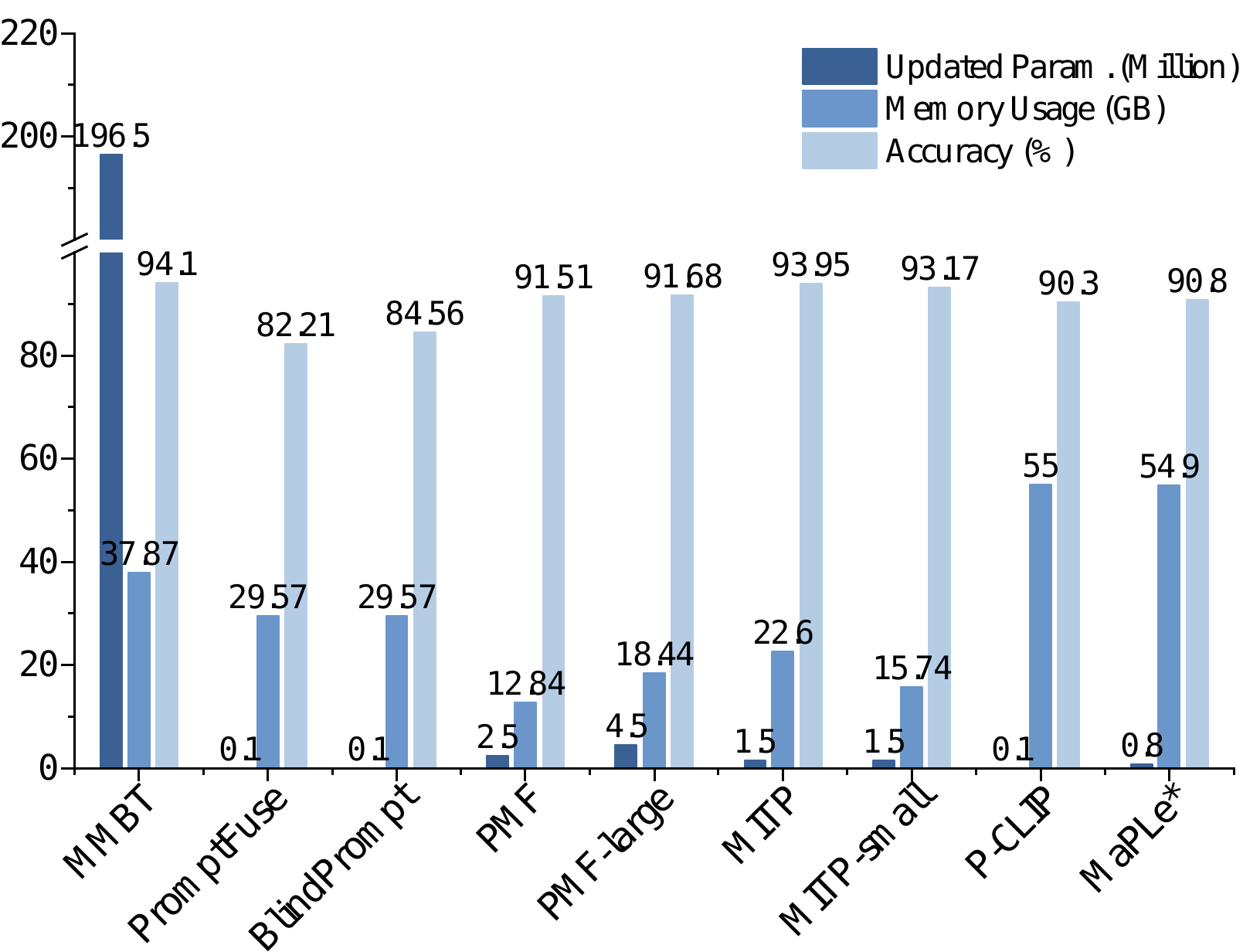}
\caption*{(a) Comparisons on overall efficiency.}
\end{minipage}
\begin{minipage}{0.34\linewidth}
\centering
\includegraphics[width=1\textwidth]{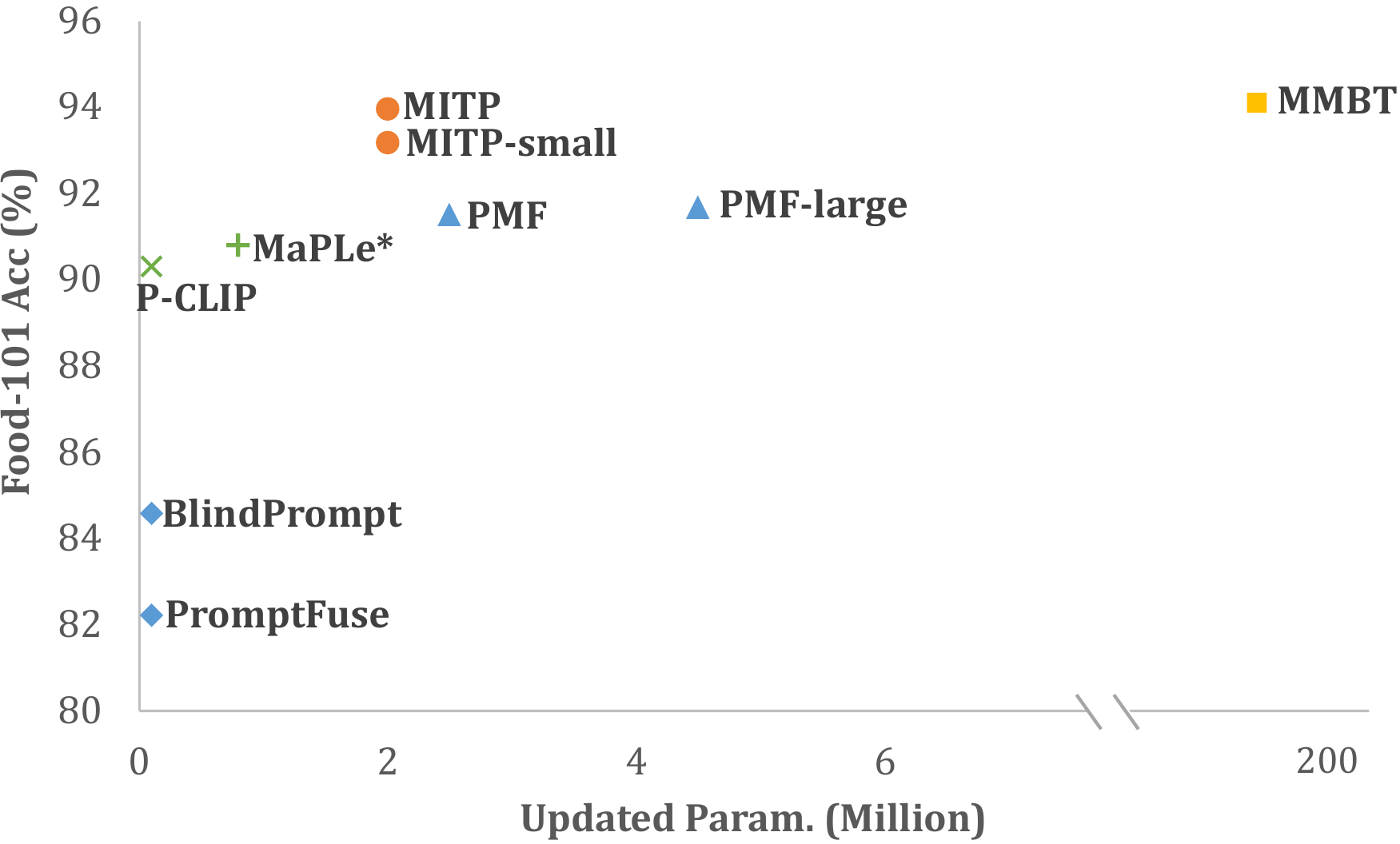}
\caption*{(b) Comparison on efficiency of trainable parameters.}
\end{minipage}
\begin{minipage}{0.34\linewidth}
\centering
\includegraphics[width=1\textwidth]{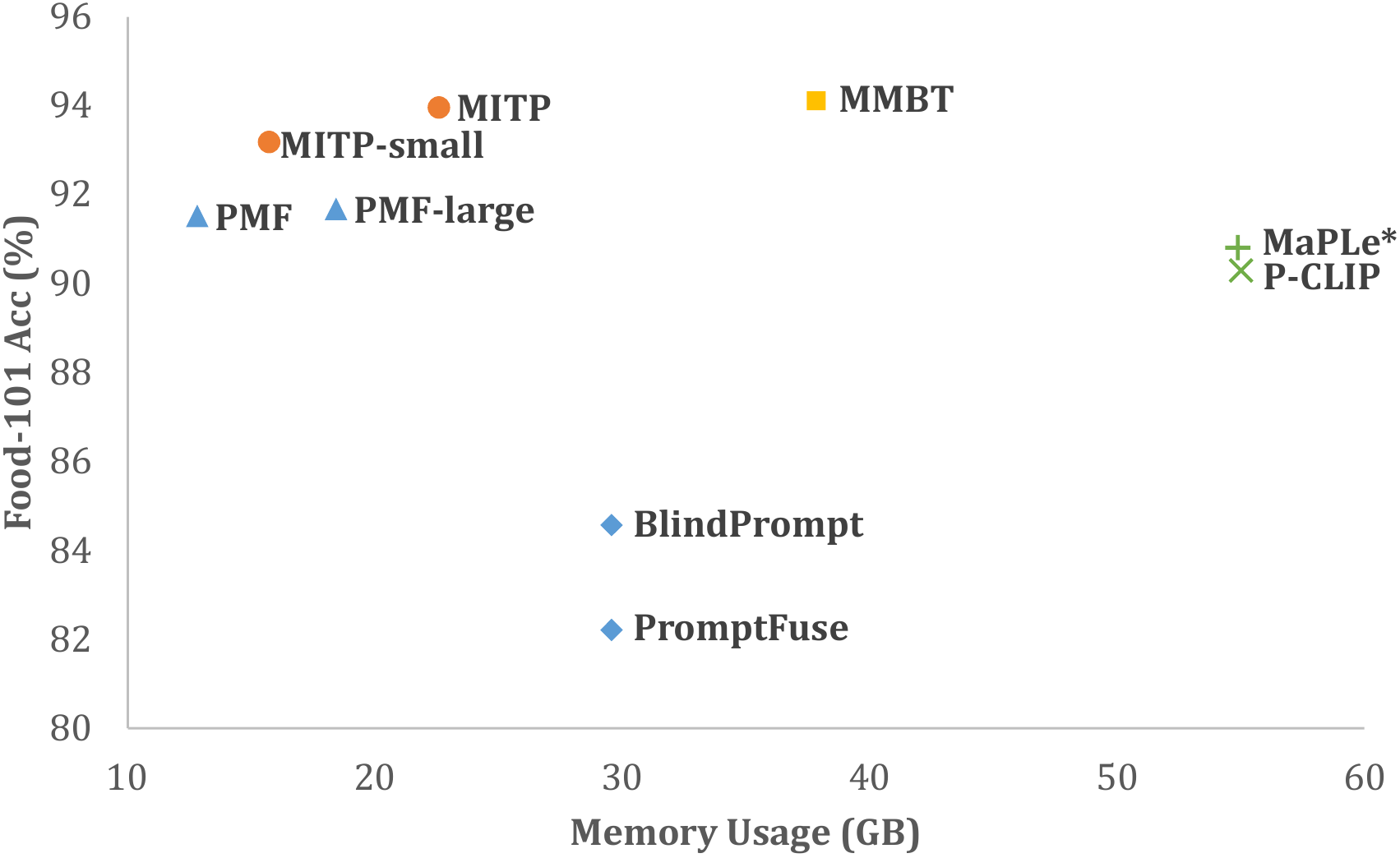}
\caption*{(c) Comparison on efficiency of memory usage in training.}
\end{minipage}
\caption{The comparison of overall efficiency among MITP and other prompt-based methods. MITP reaches high performance with only 2.0M of parameters (about 1\% of the pre-trained model) to update and relatively small memory usage in training.} 
\label{fig5}
\end{figure*}

\begin{figure*}
    \centering
    \begin{minipage}{0.34\linewidth} 
    \centering
    \includegraphics[width=1\textwidth]{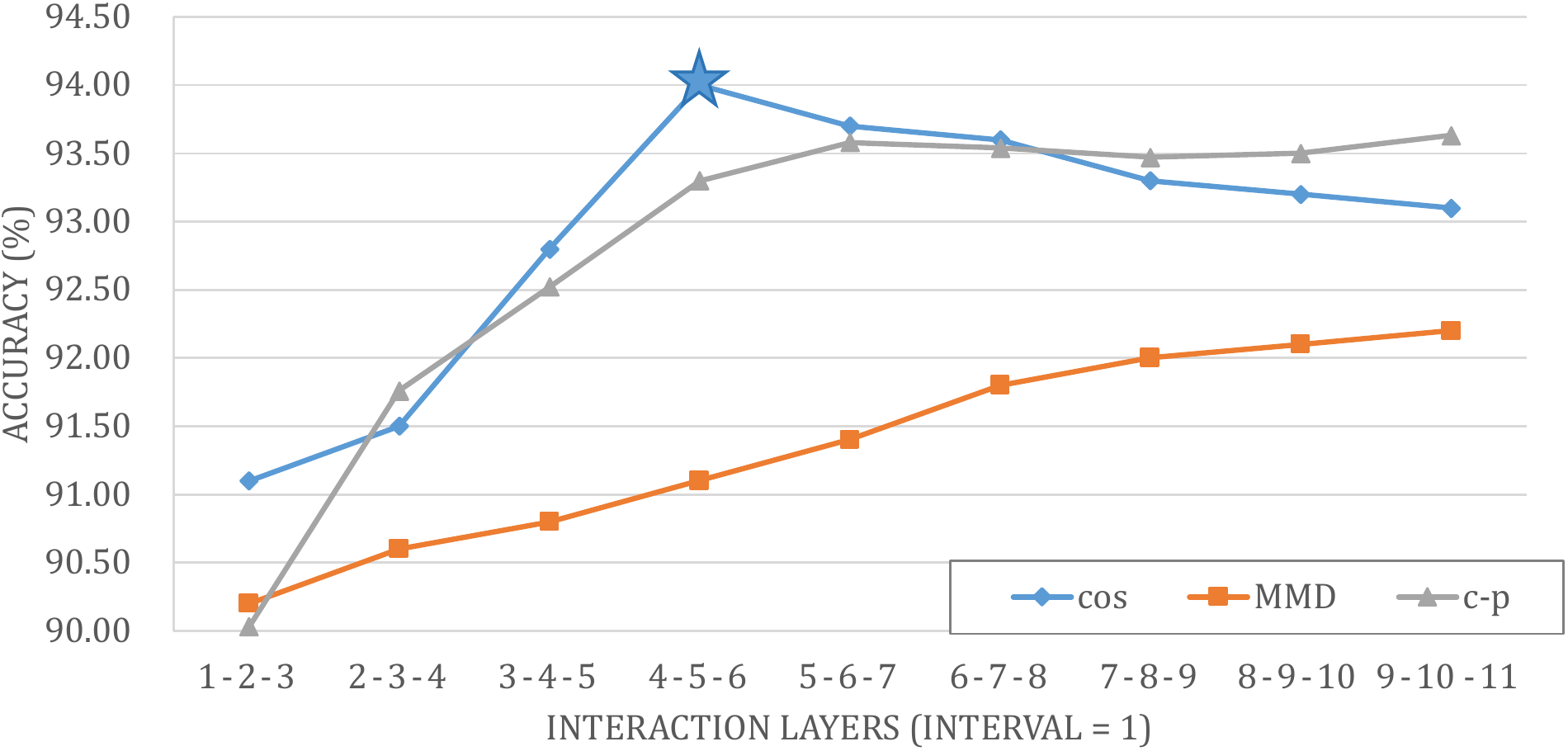}
    \caption*{(a) The results on interaction layers with interval=1.} 
    \end{minipage}
    \begin{minipage}{0.32\linewidth}
    \centering
    \includegraphics[width=1\textwidth]{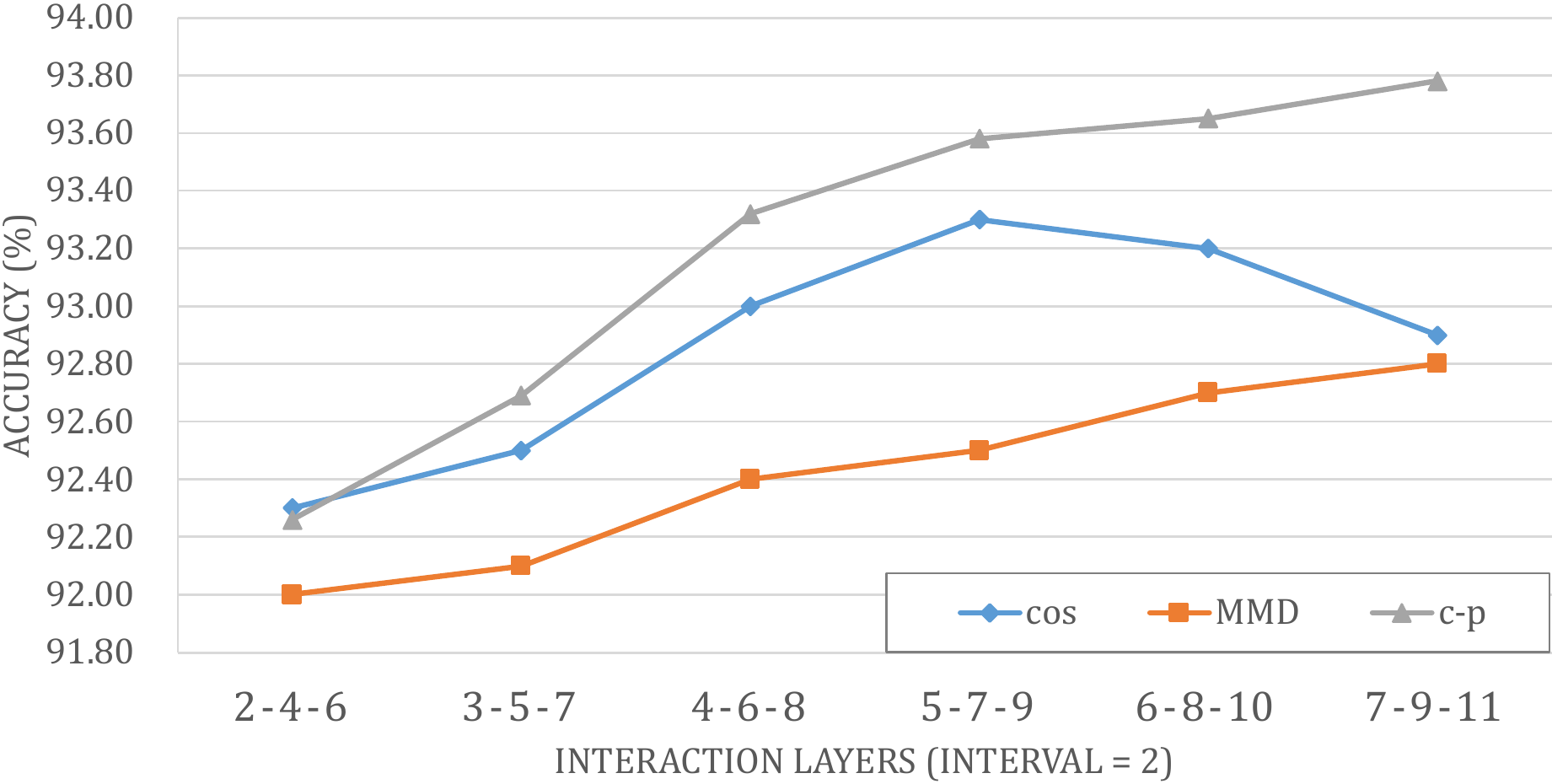}
    \caption*{(b) The results on interaction layers with interval=2.} 
    \end{minipage}
    \begin{minipage}{0.32\linewidth}
    \centering
    \includegraphics[width=1\textwidth]{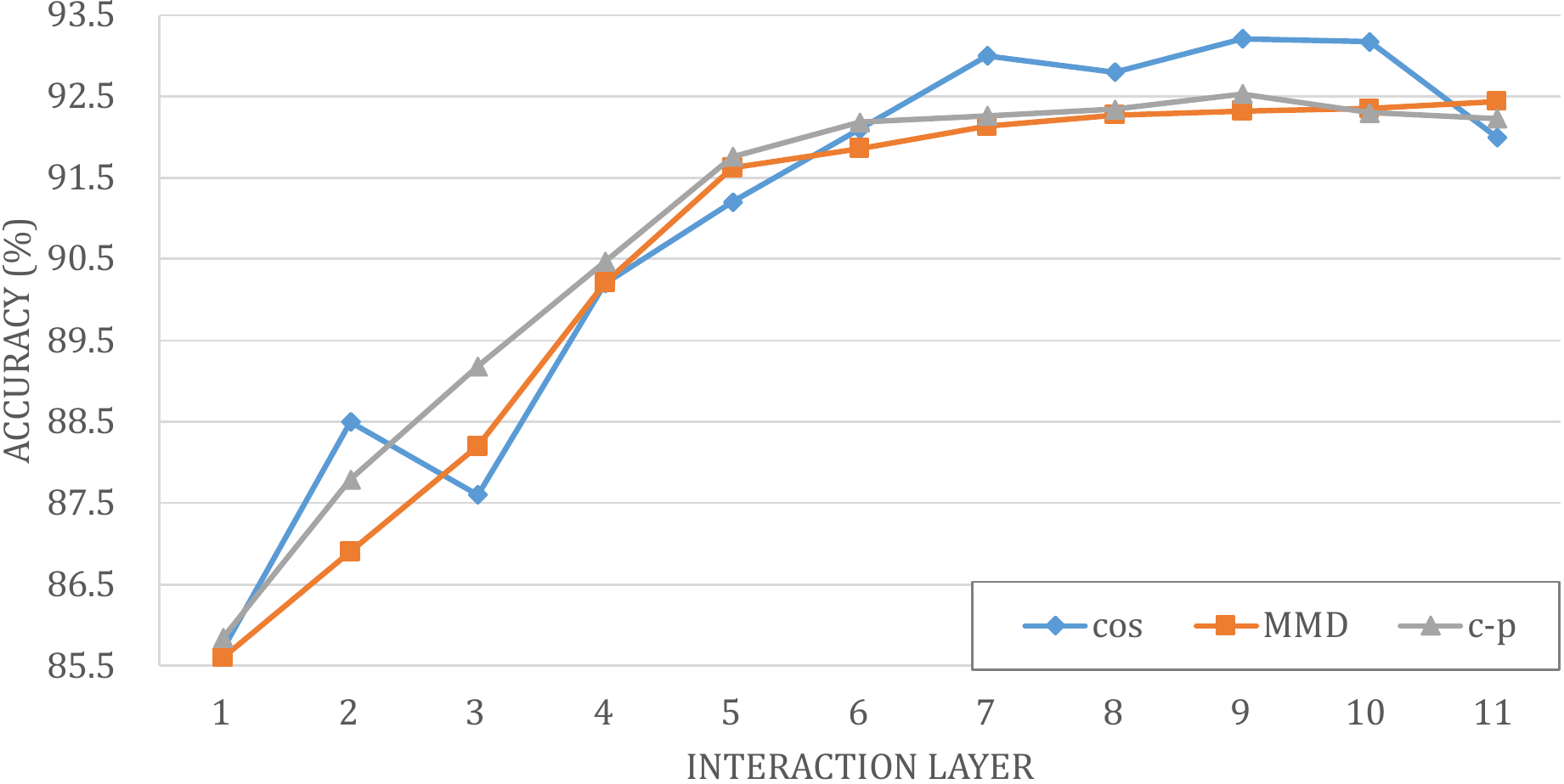}
    \caption*{(c) The results of MITP-small on a single interaction layer.} 
    \end{minipage}
    \caption{Selection of interaction layers and similarity type. In (a) and (b), the intervals of interaction layers are selected from ${1,2}$, and similarity type is selected from $\{cos, mmd,c-p\}$, where $cos$ means cosine similarity, $mmd$ means MMD similarity, and $c-p$ denotes a combination of covariance and pearson's correlation coefficient. In (c), we perform prompt interaction on one layer, with cosine similarity. Results of accuracy and memory usage is reported in (c). The results are on UPMC-Food101.}
    \label{fig8}
\end{figure*}

\subsection{Main results}

\textbf{MITP is effective on diverse image-text classification task types.} To demonstrate the effect of our method, we compare MITP with various existing late fusion strategies, prompt-tuning strategies and prompt-based interaction strategies, including CMA-CLIP~\cite{fu2022cma}, MMBT~\cite{kiela2019supervised}, PromptFuse~\cite{liang2022modular}, PMF~\cite{li2023efficient}, etc. The results of our method are shown in Table~\ref{tab:my_label1}. From the results, we observe that our proposed method achieves competitive results on all three datasets for diverse image-text task types. In detail, we can achieve an accuracy of 93.95\(\%\) on uni-label classification dataset UPMC-Food101, F1-micro/F1-macro of 65.9/56.3 on multi-label classification dataset MM-IMDB, and an accuracy of 73.45\(\%\) on multimodal visual entailment dataset SNLI-VE. The small version of our proposed method (MITP-small) also achieves competitive results on all three datasets. In particular, we outperform all existing prompt-based methods on datasets UPMC-Food101 and SNLI-VE.

\textbf{MITP is efficient in terms of trainable parameters and memory usage.} We compare the efficiency our proposed method MITP with other prompt-based methods. The efficiency is considered from perspectives of the number of trainable parameters and the maximum memory usage in training. We present comparisons of overall efficiency, efficiency of parameters, and efficiency of memory usage in Fig~\ref{fig5}. MITP achieves the second highest accuracy on UPMC-Food101 with only 2.0M parameters (about 1\(\%\) of its pre-trained foundation model) to update and a relatively small training memory usage. The illustrations show that MITP is balanced on performance and efficiency of trainable parameters and memory usage.

\subsection{Ablation study}

In this section, we conduct ablation study on UPMC-Food101. We carry out ablation study on components, to examine the effectiveness of temporal prompts, prompt interaction, and memory-inspired prompt generation strategy separately. We also illustrate illustrate which layers to perform temporal prompt interaction, and impact of different similarity types in our prompt generation strategy. Finally we discuss the robustness of the proposed method when data is insufficient, and demonstrate the effectiveness.

\textbf{Ablation study.} To verify the impact of each component in our method, we conduct component ablation study on UPMC-Food101. The results are presented in Table~\ref{tab:my_label4}. The baseline model does not introduce temporal prompts and is the same as the employed pre-trained foundation model. After leveraging prompt interaction (Model 2) and memory-inspired prompt generation strategy (Model 1) to temporal prompts (Model 3), the performance increases significantly, proving the effectiveness of our each component. The performance gap between the baseline model and other models demonstrates the importance to introduce prompts on intermediate layers, as well as the importance of prompt interaction and prompt generation strategy. 

\begin{table*}[]
    \centering
    \caption{Ablation study on components on three employed datasets. $Tmp.$ $prompt$ means temporal prompts on intermediate layers of each branch. $P. $ $interaction$ means prompt interaction through naive MLP. $Mem.$ $ strategy$ means prompt generation strategies introduced in \ref{memory_hub}.}
    \begin{tabular}{c|ccc|c|c|c}
    \toprule
    &  \begin{tabular}[c]{@{}l@{}}$Tmp.$\\ prompt\end{tabular} & \begin{tabular}[c]{@{}l@{}}$P.$\\ interaction\end{tabular} & \begin{tabular}[c]{@{}l@{}} $Mem.$\\ strategy\end{tabular} & \begin{tabular}[c]{@{}l@{}} UMPC-Food101\\Acc(\%)\end{tabular} &\begin{tabular}[c]{@{}l@{}} MM-IMDB\\F1-micro/-macro\end{tabular} &\begin{tabular}[c]{@{}l@{}} SNLI-VE\\Acc(\%)\end{tabular} \\
    \midrule
    Model 1&\checkmark & \checkmark & \checkmark & 93.95 & 65.9 / 56.3 & 73.45 \\
    Model 2&\checkmark  & \checkmark & & 92.13 & 63.5 / 54.0 & 71.89 \\
    Model 3&\checkmark & & & 90.32 & 60.9 / 51.6 & 70.65\\
    Baseline& & & & 88.80 & 60.0 / 50.8 & 68.47 \\
    \bottomrule
    \end{tabular}
    \label{tab:my_label4}
\end{table*}

\textbf{Selection of interaction layers and similarity type.} \label{interaction_layers} \label{similarity_type}
In the memory hub, various types of similarity can be chosen for prompt generation. Therefore we discuss the selection of similarity type for memory hub. We also discuss the selection of interaction layers. For the base MITP, we perform temporal prompt interaction on three layers. Empirically the intervals of selected three layers are the same, for example, (4,5,6) or (6,8,10). For the small version MITP-small, temporal prompt interaction is operated on a single layer. For similarity type, we choose cosine similarity, distribution-based similarity maximum mean discrepancy (MMD), and correlation-based combination of covariance and Pearson's correlation coefficient (denoted as covariance-pearsonr), where covariance is used to measure the similarity within temporal prompt from the same modality, and Pearson's correlation coefficient is used to measure the similarity between temporal prompts from different modalities.

We conduct experiments to explore the performance of different layers and different similarity types. The experimental results on selection of interaction layers and similarity types are shown in Fig~\ref{fig8}. The results indicate that the model reaches its highest performance at interaction layers (4,5,6) with cosine similarity. (On MM-IMDB and SNLI-VE, the best performance is achieved at interaction layers (7,8,9) with cosine similarity.) The performance of cosine similarity is generally higher than that of MMD. Cosine similarity also outperforms covariance-pearsonr when interval is 1, while covariance-pearsonr achieves higher performance when interval is 2. The reason may be that distributions differ a lot between mid-level features of different modalities. In common knowledge, deeper layers imply more semantic information, leading to better experimental results when modality interaction is conducted at deeper layers. However, this law is only supported by the model with distribution-based similarities MMD and correlation-based covariance-pearsonr. The performance of model with cosine similarity peaks when modality interaction is conducted at the mid-level layers. Interactions on shallower layers provide small improvement of performance, due to high heterogeneity between shallow-level modality features. However, interactions on shallower layers do not harm the performance. This indicates that interactions via prompts may alleviate the modality gap during interaction. As for interval of interaction layers, we can see that when cosine similarity is applied, interaction layers with interval=1 generally achieve better performance than those with interval=2; when MMD or covariance-pearsonr is applied, interaction layers with interval=2 achieve better performance. The reason may lie in the similarity-based prompt generation strategy. With cosine similarity applied, it is hard for the generated prompts to bridge the gap between layers with interval larger than 1. 

\begin{table}
    \centering
    \caption{Results on UMPC-Food101 with insufficient data. The proposed method is robust with insufficient data, and achieves an accuracy over 90\% with only 10\% of the training data used.}
    \begin{tabular}{c|c|c}
    \toprule
         \begin{tabular}[c]{@{}l@{}} Training data\\(\%)\end{tabular}& \begin{tabular}[c]{@{}l@{}} Results\\Acc(\%)\end{tabular} & \begin{tabular}[c]{@{}l@{}}Performance\\degradation(\%)\end{tabular} \\
         \midrule
         1 & 87.59 & 6.36 \\
         5 & 89.81 & 4.14 \\
         10 & 90.36 & 3.59 \\
         30 & 92.62 & 1.33 \\
         \bottomrule
    \end{tabular}
    \label{tab:my_label5}
\end{table}

\textbf{Robustness with insufficient data.} For further discussion, we demonstrate the robustness of the proposed method when the data is insufficient. We conduct experiments on UPMC-Food101 with randomly selected 1\(\%\), 5\(\%\), 10\(\%\), and 30\(\%\) of training data, the results are shown in Table~\ref{tab:my_label5}. The proposed MITP achieves an accuracy over 90$\%$ with 10$\%$ of training data. There is merely 1.33$\%$ of performance degradation when the model is trained by 30$\%$ of training data. The results indicate that our proposed method is robust with insufficient data.

\subsection{Discussions} Why is MITP able to reduce trainable parameter and memory usage? In MITP, information of extracted features is stored in temporal prompts on intermediate layers, and two-way information exchange is conducted through prompt interaction, without participation of extracted features. Prompt vectors contain much lower dimensions than extracted features, reducing spatial complexity of inter-modality information exchange. Moreover, calculative similarity-based prompt interaction ensures effective information exchange without increasing trainable parameters or model complexity. Comparing to fine-tuning methods with fine-tuning prompts leveraged on multiple layers from shallower ones to deeper ones, MITP leverages temporal prompts on selected deeper layers, significantly reduces the memory usage.

\section{Conclusions}

In this paper, we propose a memory-inspired temporal prompt interaction method for image-text classification. Our approach interacts temporal prompts on intermediate layers of the LMM foundation model to imitate human memory. Sufficient inter-modality interactions are leveraged through direct prompt interactions, while the number of parameters to update and the memory usage in training are compressed. The results can demonstrate our effectiveness. In particular, we achieve the best results among prompt-based interaction methods and outperforms prompt-tuning methods with the same LMM foundation model. However, a limitation of our proposed method is that temporal prompts are added in prefix of input tokens, so that the knowledge inside the foundation model is not fully leveraged. In the future, we plan to design prompt strategy to conduct more sufficient interaction with the input tokens.

{
    \small
    \bibliographystyle{ieeenat_fullname}
    \bibliography{main}

\begin{thebibliography}{34}
\providecommand{\natexlab}[1]{#1}
\providecommand{\url}[1]{\texttt{#1}}
\expandafter\ifx\csname urlstyle\endcsname\relax
  \providecommand{\doi}[1]{doi: #1}\else
  \providecommand{\doi}{doi: \begingroup \urlstyle{rm}\Url}\fi

\bibitem[Arevalo et~al.(2017)Arevalo, Solorio, Montes-y G{\'o}mez, and Gonz{\'a}lez]{arevalo2017gated}
John Arevalo, Thamar Solorio, Manuel Montes-y G{\'o}mez, and Fabio~A Gonz{\'a}lez.
\newblock Gated multimodal units for information fusion.
\newblock \emph{arXiv preprint arXiv:1702.01992}, 2017.

\bibitem[Baddeley(2010)]{baddeley2010working}
Alan Baddeley.
\newblock Working memory.
\newblock \emph{Current biology}, 20\penalty0 (4):\penalty0 R136--R140, 2010.

\bibitem[Deng et~al.(2021)Deng, Kang, Yang, Hao, Li, and Liu]{deng2021dense}
Huan Deng, Peipei Kang, Zhenguo Yang, Tianyong Hao, Qing Li, and Wenyin Liu.
\newblock Dense fusion network with multimodal residual for sentiment classification.
\newblock In \emph{2021 IEEE International Conference on Multimedia and Expo (ICME)}, pages 1--6. IEEE, 2021.

\bibitem[Doumas et~al.(2022)Doumas, Puebla, Martin, and Hummel]{doumas2022theory}
Leonidas~AA Doumas, Guillermo Puebla, Andrea~E Martin, and John~E Hummel.
\newblock A theory of relation learning and cross-domain generalization.
\newblock \emph{Psychological review}, 2022.

\bibitem[Fu et~al.(2022)Fu, Xu, Liu, Liu, Xie, Wang, Liu, Sun, and Wang]{fu2022cma}
Jinmiao Fu, Shaoyuan Xu, Huidong Liu, Yang Liu, Ning Xie, Chien-Chih Wang, Jia Liu, Yi Sun, and Bryan Wang.
\newblock Cma-clip: Cross-modality attention clip for text-image classification.
\newblock In \emph{2022 IEEE International Conference on Image Processing (ICIP)}, pages 2846--2850. IEEE, 2022.

\bibitem[Hazarika et~al.(2020)Hazarika, Zimmermann, and Poria]{hazarika2020misa}
Devamanyu Hazarika, Roger Zimmermann, and Soujanya Poria.
\newblock Misa: Modality-invariant and-specific representations for multimodal sentiment analysis.
\newblock In \emph{Proceedings of the 28th ACM international conference on multimedia}, pages 1122--1131, 2020.

\bibitem[Hu and Singh(2021)]{hu2021unit}
Ronghang Hu and Amanpreet Singh.
\newblock Unit: Multimodal multitask learning with a unified transformer.
\newblock In \emph{Proceedings of the IEEE/CVF International Conference on Computer Vision}, pages 1439--1449, 2021.

\bibitem[Huang et~al.(2020)Huang, Zeng, Liu, Fu, and Fu]{huang2020pixel}
Zhicheng Huang, Zhaoyang Zeng, Bei Liu, Dongmei Fu, and Jianlong Fu.
\newblock Pixel-bert: Aligning image pixels with text by deep multi-modal transformers.
\newblock \emph{arXiv preprint arXiv:2004.00849}, 2020.

\bibitem[Jia et~al.(2022)Jia, Tang, Chen, Cardie, Belongie, Hariharan, and Lim]{jia2022visual}
Menglin Jia, Luming Tang, Bor-Chun Chen, Claire Cardie, Serge Belongie, Bharath Hariharan, and Ser-Nam Lim.
\newblock Visual prompt tuning.
\newblock In \emph{European Conference on Computer Vision}, pages 709--727. Springer, 2022.

\bibitem[Khattak et~al.(2023)Khattak, Rasheed, Maaz, Khan, and Khan]{khattak2023maple}
Muhammad~Uzair Khattak, Hanoona Rasheed, Muhammad Maaz, Salman Khan, and Fahad~Shahbaz Khan.
\newblock Maple: Multi-modal prompt learning.
\newblock In \emph{Proceedings of the IEEE/CVF Conference on Computer Vision and Pattern Recognition}, pages 19113--19122, 2023.

\bibitem[Kiela et~al.(2019)Kiela, Bhooshan, Firooz, Perez, and Testuggine]{kiela2019supervised}
Douwe Kiela, Suvrat Bhooshan, Hamed Firooz, Ethan Perez, and Davide Testuggine.
\newblock Supervised multimodal bitransformers for classifying images and text.
\newblock \emph{arXiv preprint arXiv:1909.02950}, 2019.

\bibitem[Kim et~al.(2021)Kim, Son, and Kim]{pmlr-v139-kim21k}
Wonjae Kim, Bokyung Son, and Ildoo Kim.
\newblock Vilt: Vision-and-language transformer without convolution or region supervision.
\newblock In \emph{Proceedings of the 38th International Conference on Machine Learning}, pages 5583--5594. PMLR, 2021.

\bibitem[Li et~al.()Li, Yatskar, Yin, Hsieh, and Chang]{li1908visualbert}
Liunian~Harold Li, Mark Yatskar, D Yin, CJ Hsieh, and KW Chang.
\newblock Visualbert: A simple and performant baseline for vision and language. arxiv 2019.
\newblock \emph{arXiv preprint arXiv:1908.03557}.

\bibitem[Li et~al.(2023)Li, Quan, Zhu, and Yang]{li2023efficient}
Yaowei Li, Ruijie Quan, Linchao Zhu, and Yi Yang.
\newblock Efficient multimodal fusion via interactive prompting.
\newblock In \emph{Proceedings of the IEEE/CVF Conference on Computer Vision and Pattern Recognition}, pages 2604--2613, 2023.

\bibitem[Li et~al.(2022)Li, Xu, Zhu, and Zhao]{li2022clmlf}
Zhen Li, Bing Xu, Conghui Zhu, and Tiejun Zhao.
\newblock Clmlf: a contrastive learning and multi-layer fusion method for multimodal sentiment detection.
\newblock \emph{arXiv preprint arXiv:2204.05515}, 2022.

\bibitem[Liang et~al.(2021)Liang, Lyu, Fan, Wu, Cheng, Wu, Chen, Wu, Lee, Zhu, et~al.]{liang2021multibench}
Paul~Pu Liang, Yiwei Lyu, Xiang Fan, Zetian Wu, Yun Cheng, Jason Wu, Leslie Chen, Peter Wu, Michelle~A Lee, Yuke Zhu, et~al.
\newblock Multibench: Multiscale benchmarks for multimodal representation learning.
\newblock \emph{arXiv preprint arXiv:2107.07502}, 2021.

\bibitem[Liang et~al.(2022)Liang, Zhao, and Sch{\"u}tze]{liang2022modular}
Sheng Liang, Mengjie Zhao, and Hinrich Sch{\"u}tze.
\newblock Modular and parameter-efficient multimodal fusion with prompting.
\newblock \emph{arXiv preprint arXiv:2203.08055}, 2022.

\bibitem[Liu et~al.(2023)Liu, Zheng, Du, Ding, Qian, Yang, and Tang]{liu2023gpt}
Xiao Liu, Yanan Zheng, Zhengxiao Du, Ming Ding, Yujie Qian, Zhilin Yang, and Jie Tang.
\newblock Gpt understands, too.
\newblock \emph{AI Open}, 2023.

\bibitem[Liu et~al.(2021)Liu, Fan, Zhang, Dong, Funkhouser, and Yi]{liu2021contrastive}
Yunze Liu, Qingnan Fan, Shanghang Zhang, Hao Dong, Thomas Funkhouser, and Li Yi.
\newblock Contrastive multimodal fusion with tupleinfonce.
\newblock In \emph{Proceedings of the IEEE/CVF International Conference on Computer Vision}, pages 754--763, 2021.

\bibitem[Lu et~al.(2021)Lu, Zhang, Huang, Wang, Yu, Zhao, and Wu]{lu2021future}
Yujie Lu, Shengyu Zhang, Yingxuan Huang, Luyao Wang, Xinyao Yu, Zhou Zhao, and Fei Wu.
\newblock Future-aware diverse trends framework for recommendation.
\newblock In \emph{Proceedings of the Web Conference 2021}, pages 2992--3001, 2021.

\bibitem[Mai et~al.(2022)Mai, Zeng, Zheng, and Hu]{mai2022hybrid}
Sijie Mai, Ying Zeng, Shuangjia Zheng, and Haifeng Hu.
\newblock Hybrid contrastive learning of tri-modal representation for multimodal sentiment analysis.
\newblock \emph{IEEE Transactions on Affective Computing}, 2022.

\bibitem[Nagrani et~al.(2021)Nagrani, Yang, Arnab, Jansen, Schmid, and Sun]{nagrani2021attention}
Arsha Nagrani, Shan Yang, Anurag Arnab, Aren Jansen, Cordelia Schmid, and Chen Sun.
\newblock Attention bottlenecks for multimodal fusion.
\newblock \emph{Advances in Neural Information Processing Systems}, 34:\penalty0 14200--14213, 2021.

\bibitem[Narayana et~al.(2019)Narayana, Pednekar, Krishnamoorthy, Sone, and Basu]{narayana2019huse}
Pradyumna Narayana, Aniket Pednekar, Abishek Krishnamoorthy, Kazoo Sone, and Sugato Basu.
\newblock Huse: Hierarchical universal semantic embeddings.
\newblock \emph{arXiv preprint arXiv:1911.05978}, 2019.

\bibitem[Radford et~al.(2021)Radford, Kim, Hallacy, Ramesh, Goh, Agarwal, Sastry, Askell, Mishkin, Clark, et~al.]{radford2021learning}
Alec Radford, Jong~Wook Kim, Chris Hallacy, Aditya Ramesh, Gabriel Goh, Sandhini Agarwal, Girish Sastry, Amanda Askell, Pamela Mishkin, Jack Clark, et~al.
\newblock Learning transferable visual models from natural language supervision.
\newblock In \emph{International conference on machine learning}, pages 8748--8763. PMLR, 2021.

\bibitem[Sun et~al.(2022)Sun, Wang, Liu, Chen, and Lin]{sun2022cubemlp}
Hao Sun, Hongyi Wang, Jiaqing Liu, Yen-Wei Chen, and Lanfen Lin.
\newblock Cubemlp: An mlp-based model for multimodal sentiment analysis and depression estimation.
\newblock In \emph{Proceedings of the 30th ACM International Conference on Multimedia}, pages 3722--3729, 2022.

\bibitem[Vielzeuf et~al.(2018)Vielzeuf, Lechervy, Pateux, and Jurie]{vielzeuf2018centralnet}
Valentin Vielzeuf, Alexis Lechervy, St{\'e}phane Pateux, and Fr{\'e}d{\'e}ric Jurie.
\newblock Centralnet: a multilayer approach for multimodal fusion.
\newblock In \emph{Proceedings of the European Conference on Computer Vision (ECCV) Workshops}, pages 0--0, 2018.

\bibitem[Wang et~al.(2015)Wang, Kumar, Thome, Cord, and Precioso]{wang2015recipe}
Xin Wang, Devinder Kumar, Nicolas Thome, Matthieu Cord, and Frederic Precioso.
\newblock Recipe recognition with large multimodal food dataset.
\newblock In \emph{2015 IEEE International Conference on Multimedia \& Expo Workshops (ICMEW)}, pages 1--6. IEEE, 2015.

\bibitem[Xie et~al.(2018)Xie, Lai, Doran, and Kadav]{xie2018visual}
Ning Xie, Farley Lai, Derek Doran, and Asim Kadav.
\newblock Visual entailment task for visually-grounded language learning.
\newblock \emph{arXiv preprint arXiv:1811.10582}, 2018.

\bibitem[Xie et~al.(2019)Xie, Lai, Doran, and Kadav]{xie2019visual}
Ning Xie, Farley Lai, Derek Doran, and Asim Kadav.
\newblock Visual entailment: A novel task for fine-grained image understanding.
\newblock \emph{arXiv preprint arXiv:1901.06706}, 2019.

\bibitem[Xue and Marculescu(2023)]{xue2023dynamic}
Zihui Xue and Radu Marculescu.
\newblock Dynamic multimodal fusion.
\newblock In \emph{Proceedings of the IEEE/CVF Conference on Computer Vision and Pattern Recognition}, pages 2574--2583, 2023.

\bibitem[Zadeh et~al.(2017)Zadeh, Chen, Poria, Cambria, and Morency]{zadeh2017tensor}
Amir Zadeh, Minghai Chen, Soujanya Poria, Erik Cambria, and Louis-Philippe Morency.
\newblock Tensor fusion network for multimodal sentiment analysis.
\newblock \emph{arXiv preprint arXiv:1707.07250}, 2017.

\bibitem[Zadeh et~al.(2018)Zadeh, Liang, Mazumder, Poria, Cambria, and Morency]{zadeh2018memory}
Amir Zadeh, Paul~Pu Liang, Navonil Mazumder, Soujanya Poria, Erik Cambria, and Louis-Philippe Morency.
\newblock Memory fusion network for multi-view sequential learning.
\newblock In \emph{Proceedings of the AAAI conference on artificial intelligence}, 2018.

\bibitem[Zhou et~al.(2022{\natexlab{a}})Zhou, Yang, Loy, and Liu]{zhou2022conditional}
Kaiyang Zhou, Jingkang Yang, Chen~Change Loy, and Ziwei Liu.
\newblock Conditional prompt learning for vision-language models.
\newblock In \emph{Proceedings of the IEEE/CVF Conference on Computer Vision and Pattern Recognition}, pages 16816--16825, 2022{\natexlab{a}}.

\bibitem[Zhou et~al.(2022{\natexlab{b}})Zhou, Yang, Loy, and Liu]{zhou2022learning}
Kaiyang Zhou, Jingkang Yang, Chen~Change Loy, and Ziwei Liu.
\newblock Learning to prompt for vision-language models.
\newblock \emph{International Journal of Computer Vision}, 130\penalty0 (9):\penalty0 2337--2348, 2022{\natexlab{b}}.

\end{thebibliography}
}


\end{document}